# Back To The Future: A Hybrid Transformer-XGBoost Model for Action-oriented Future-proofing Nowcasting

Ziheng Sun

George Mason University

zsun@gmu.edu

**Abstract**:

Inspired by the iconic movie *Back to the Future*, this paper explores an innovative adaptive nowcasting approach that reimagines the relationship between present actions and future outcomes. In the movie, characters travel through time to manipulate past events, aiming to create a better future. Analogously, our framework employs predictive insights about the future to inform and adjust present conditions. This dual-stage model integrates the forecasting power of Transformers (future visionary) with the interpretability and efficiency of XGBoost (decision maker), enabling a seamless loop of future prediction and present adaptation. Through experimentation with meteorological datasets, we demonstrate the framework's advantage in achieving more accurate forecasting while guiding actionable interventions for real-time applications.

**Keywords**:

Hybrid Modeling; Time-Series Forecasting; Real-Time Decision-Making; Transformer-XGBoost Framework; Dynamic Adaptation

## 1. Introduction

The interplay between past, present, and future is a central theme in the iconic movie Back to the Future, where small alterations in past events have profound, cascading effects on the future [1]. This concept mirrors the intricate and often non-linear relationships in real-world systems, where predictions about the future are not merely passive observations but active drivers of current decisions and behaviors. In the film, the characters reshape their present and future by altering past events, reflecting the power of temporal causality—the idea that events in time are interconnected, and that actions taken now have consequences for the future. In much the same way, effective nowcasting—predicting short-term outcomes like weather, natural hazards, health events, or traffic patterns—should not only anticipate what will happen but also incorporate how those predictions can influence present decisions and conditions.

Traditional nowcasting methods, however, often focus exclusively on making predictions about future states without considering the active feedback loop that can be created by those predictions [2]. They treat future events as isolated targets, independent of the present context in which decisions are being made [3–5]. In meteorology, for instance, a model may predict rainfall or temperature for the next few hours, but it often doesn't incorporate how current actions, based on these predictions, could influence future conditions. For example, adjusting traffic patterns to prevent flooding or reallocating emergency services to areas at risk could have a significant impact on the outcomes. Ideally, forecasting should not only predict future weather events but also consider how the interventions informed by these predictions could affect future conditions, allowing for dynamic adjustments that improve both immediate and long-term decision-making. This limitation also extends to other dynamic systems like healthcare and traffic management, where predictions—though accurate—are often disconnected from immediate actions that could alter the course of the predicted event.

We propose a "back to the future" (BTTF) strategy to nowcasting, which combines forecasting and real-time decision-making into a continuous, adaptive feedback loop. Drawing inspiration from Back to the Future, this hybrid approach envisions a system where predictions are not merely used to anticipate future outcomes but are actively employed to optimize present conditions. In the framework, the future is not a static endpoint; it is a dynamic force that actively shapes the present. For example, in the context of meteorology, a Transformer model [6] can forecast the weather in the coming hours or days, while simultaneously, an optimization model like XGBoost [7] can use that forecast to adjust current parameters—such as adjusting traffic flows in anticipation of a storm or reallocating healthcare resources based on predicted health risks. This approach brings the past, present, and future into a unified, dynamic decision-making process.

In the context of AI-driven nowcasting [8,9], the combination of these two models—Transformers for prediction and XGBoost for optimization—offers a powerful means of achieving more than just accurate forecasts. By leveraging the predictive power of Transformers to capture complex temporal dependencies and using XGBoost's interpretability and efficiency to influence real-time decisions, the system can adapt based on evolving conditions. This feedback loop is essential in dynamic systems where rapid adjustments are necessary to address unforeseen changes. For example, in traffic management, knowing that a heavy storm is predicted could allow for real-time traffic rerouting, reducing congestion and minimizing accidents before they occur. In healthcare, anticipating a spike in respiratory diseases due to changing weather conditions can enable preemptive allocation of medical resources, potentially saving lives.

This BTTF strategy is expected to potentially redefine nowcasting as a proactive, dynamic process that moves beyond traditional forecasting. Instead of merely predicting what will happen, the model actively reshapes the present state to optimize future outcomes. However, creating such a system requires bridging the gap between forecasting and adaptive decision-making, which has traditionally been difficult to achieve. The challenge lies in integrating models that can capture complex, non-linear relationships, such as those provided by deep learning models like Transformers, with models that offer real-time efficiency and interpretability, like XGBoost. By combining these models, we can achieve a more holistic approach to nowcasting that not only predicts the future but also actively influences and improves present conditions to ensure better outcomes.

In this paper, we present a concept framework that integrates these two paradigms—forecasting and optimization—through a feedback loop that draws on the strengths of both Transformers and XGBoost. Our approach aims to push the boundaries of what is possible in nowcasting, providing superior accuracy, adaptability, and actionable insights. The results from comprehensive evaluations using meteorological datasets demonstrate that this hybrid approach significantly enhances forecasting performance while providing a more proactive, adaptive decision-making process. By rethinking the role of predictions in dynamic systems, our method introduces a new paradigm for nowcasting that reflects the interconnectedness of the past, present, and future, just like Back to the Future does with time.

## 2. Related work

In recent years, the adoption of self-attention mechanisms and Transformer models has revolutionized the field of time series forecasting, especially in nowcasting, where the need for accurate and timely predictions is critical. Traditional methods, such as XGBoost [7] and Random Forest [10], excelled in simpler, linear scenarios but struggled with the complexity of capturing non-linear dependencies inherent in meteorological data. While machine learning techniques like LSTMs [11] and CNNs [12] have enabled modeling of temporal and spatial relationships, their limitations in processing long-range dependencies and handling high-dimensional data have become apparent. The Transformer model, originally designed for natural language processing tasks, addresses these shortcomings with its self-attention mechanism, which excels in capturing long-range dependencies across sequences [6]. Its capacity to process data in parallel and weigh the importance of different parts of an input sequence, regardless of temporal or spatial distance, has led to its successful application in weather forecasting, particularly in tasks like precipitation prediction and atmospheric motion modeling [13–15]. Studies have highlighted the model's ability to efficiently process large-scale data, such as satellite imagery and weather sensor outputs, making it highly suitable for real-time forecasting environments [16,17].

Despite its strengths, Transformer models face significant challenges, primarily due to their high computational cost, especially in terms of memory and processing power. The resource demands can hinder their deployment in real-time nowcasting systems, where both speed and accuracy are crucial. Meanwhile, their "black-box" nature complicates interpretability, making it difficult for users to understand how input features influence predictions, which is a critical factor in operational forecasting environments[18]. To overcome these issues, recent studies have explored hybrid approaches that combine the capabilities of Transformer models with more interpretable techniques, such as gradient boosting models like XGBoost. These hybrids aim to leverage the Transformer's strength in capturing long-term dependencies while incorporating the efficiency and transparency of gradient boosting [19]. However, challenges remain in achieving effective integration, particularly in adapting to real-time data while maintaining both accuracy and speed. The complexity of hybrid models can lead to overfitting, reducing their ability to generalize across new datasets or dynamic scenarios. Our research seeks to address these challenges by integrating Transformer-based forecasting with XGBoost for real-time adaptation, offering a more interpretable, efficient, and adaptive solution for weather prediction in fast-changing environments.

## 3. Methodology

The Back to the Future (BTTF) framework (as shown in Fig. 1) is designed to address two essential aspects of real-time decision-making systems: future prediction and present adaptation. By splitting these tasks into two distinct modules, the Future Visionary and the Decision Maker, ensures a dynamic, iterative process that combines forecasting with actionable interventions.

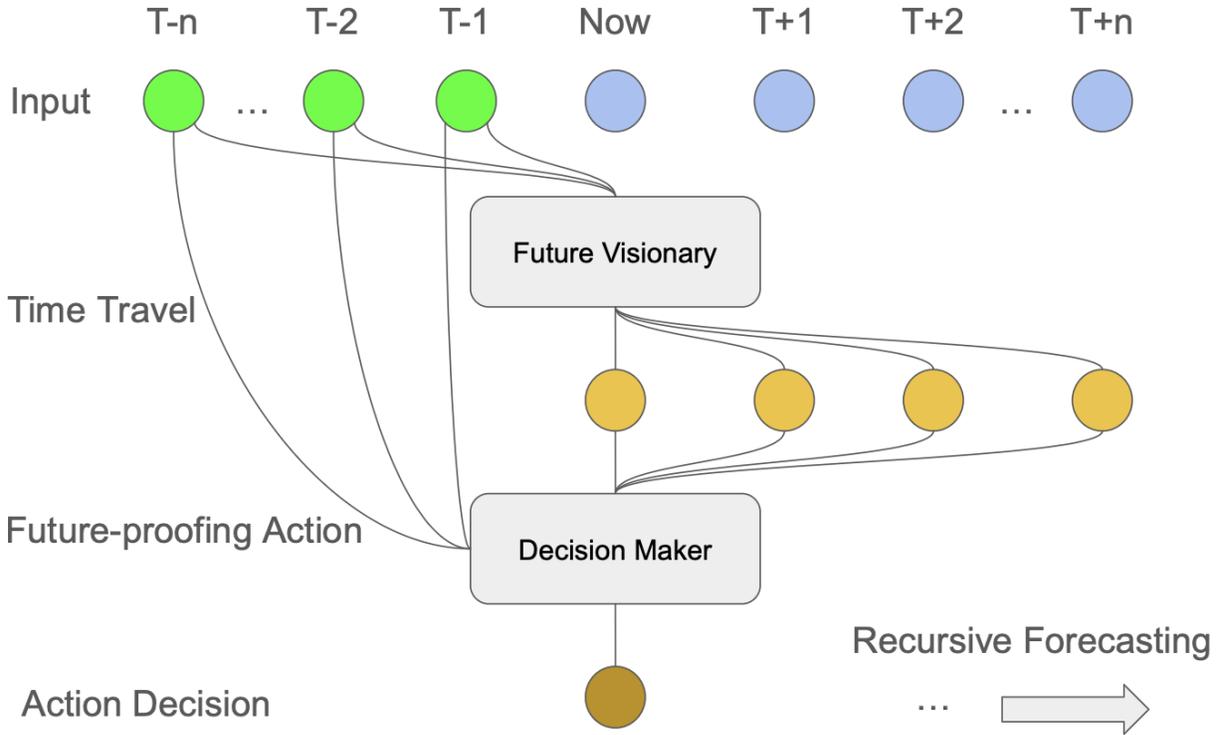

Figure 1. The concept diagram of "Back to the Future" Nowcasting Framework

## *3.1 Future Visionary for Prediction*

The Future Visionary module focuses on accurately predicting future states of the system based on historical data. While the Transformer model, with its state-of-the-art sequence-to-sequence (seq2seq) capabilities [20], is a natural choice for this task, the framework is flexible and can incorporate other seq2seq models like LSTM, or graphical neural network (GNN) [21,22]. We select it due to its proven performance in handling complex, non-linear temporal dependencies. The Transformer utilizes a multi-head self-attention mechanism to process the input sequence and predict future states $Y_i$, where $Y_i$ represents the forecasted values for the forecast horizon hh. The self-attention mechanism assigns dynamic weights to each time step in the sequence, allowing the model to capture long-range dependencies and intricate temporal patterns. Positional encodings are incorporated to maintain the sequential order of the data, ensuring that the temporal relationships between observations are preserved. The Transformer's ability to focus on relevant past data at each time step makes it highly effective in modeling sequences with complex, non-linear relationships.

The objective function for the visionary model is the same as most seq2seq models, minimizing the mean squared error (MSE) between the predicted and actual values over the forecast horizon:

$$E_{visionary} = \frac{1}{N} \sum_{i=1}^{N} (\widehat{Y_i} - Y_i)^2$$

where N is the number of the forecasted samples in each step. This loss function penalizes large deviations between predictions and actual observations, driving the model to produce accurate forecasts. The architecture, with its attention-based layers, is particularly well-suited for handling the complex, non-linear temporal patterns that are often present in meteorological data. Users also could change the MSE to another loss function regarding various data contexts and patterns [23].

## 3.2 Decision Maker for Present Adaptation

The Decision Maker module is to use the future predictions $Y_t$ generated by the visionary model to optimize and adapt the present state $X_t$. This process is essential for implementing real-time interventions that bridge the gap between prediction and action. In this study, XGBoost is employed to perform this task to handle complex relationships and provide interpretable results. Once the visionary model generates future predictions $\hat{Y}_t$, these forecasts are fed into the XGBoost model, which is used to determine adjustments for the present state $X_t$. The updated state is expressed as:

$$X_{adjusted} = X_t + \Delta X_t$$

where $\Delta X_t$ represents the output future time series by the Transformer model needed to optimize the current state. The XGBoost model performs this adaptation by minimizing a loss function that incorporates both the prediction accuracy and model complexity. The objective function is defined as:

$$L_{XGBoost} = \frac{1}{N}\sum_{i=1}^{N}(\hat{X}_\iota - X_i)^2 + \lambda \sum_{j=1}^{M}|w_j|$$

where $\hat{X}_\iota$ is the predicted adjusted state, $X_i$ is the actual present state, and $\lambda$ is a regularization term that penalizes overfitting by constraining the complexity of the model. The regularization term $\sum_{j=1}^{M}|w_j|$ ensures that the model remains interpretable by preventing overly complex decision trees.

XGBoost's ability to rank feature importance provides interpretability, enabling stakeholders to understand which factors are driving the suggested adjustments. For instance, features such as wind speed variability or humidity fluctuations may emerge as key contributors to short-term interventions, guiding targeted actions to improve current conditions. This interpretability is essential for real-time decision-making applications, where transparency in the model's reasoning is crucial.

## 3.3 Integrated Framework

The hybrid Transformer-XGBoost framework operates as a feedback loop that alternates between forecasting and adaptation. In the first stage, the Transformer model predicts future states, capturing the temporal dependencies and trends within the dataset. In the second stage, XGBoost uses these predictions to compute optimized adjustments $\Delta X_t$ enabling actionable interventions in the present state. This iterative process allows for continuous updates, with the system refining its predictions and adjustments based on the latest available data. By combining the Transformer's predictive accuracy with XGBoost's stable adaptive capabilities, the framework aims to ensure that immediate decisions are informed by robust forecasts. The integration of the two components is expected to create a more unified system that is both forward-looking and reactive, addressing the dual demands of accuracy and real-time applicability. This architecture is particularly well-suited for environments where the ability to influence outcomes depends on understanding future trends and responding proactively to changing conditions.

## 4. Experiment and Results

To demonstrate the advantages of the BTTF framework over standalone models in conventional time series forecasting, we conducted a series of experiments.

*4.1 Data Preparation and Features*

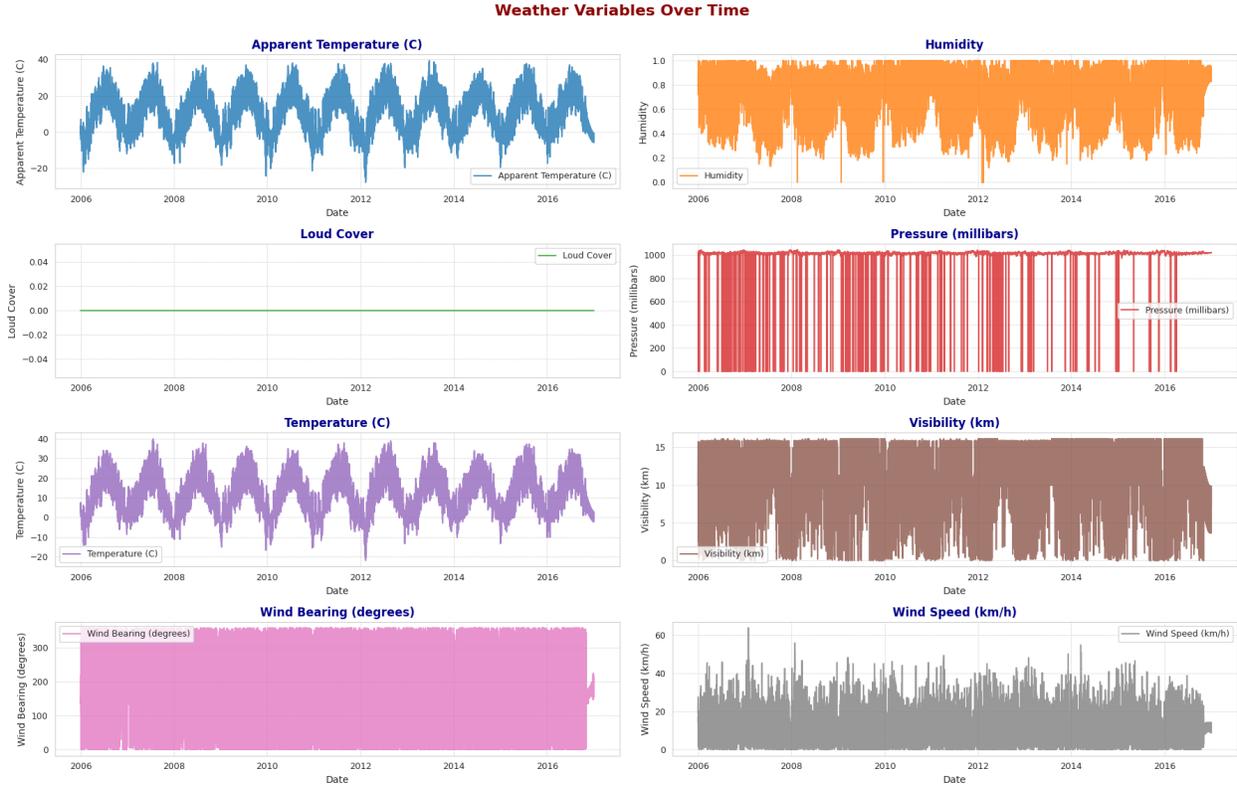

Figure 1. Used weather data statistics [24]

We downloaded a dataset from Kaggle [24] (Fig. 1) to test if this new framework can effectively forecast. The dataset has eight meteorological variables: Temperature (°C), Apparent Temperature (°C), Humidity, Wind Speed (km/h), Wind Bearing (degrees), Visibility (km), Loud Cover, and Pressure (millibars). These variables are recorded over time to capture the dynamic nature of weather conditions. The cleaned dataset consists of over 96,440 observations, which have been preprocessed to exclude missing values and irrelevant columns. This ensures a robust foundation for sequence-based forecasting models. To prepare the data for the hybrid model, sequences are constructed to capture temporal dynamics. Let $X_t$ represent the observed variables at time t, and the input sequence for the model is defined as:

$$X_{input} = \{X_{t-k}, X_{t-k+1}, \ldots, X_{t-1}\}$$

where k is the sequence length, representing the historical context that informs the prediction at time $t$. Each sequence is normalized to stabilize numerical computations and ensure consistency across variables, preventing scale disparities from impacting model performance. This step is essential to ensure that different units of measurement (e.g., temperature in Celsius, wind speed in km/h) do not introduce bias in the learning process.

In future, we could enhance this data with derived features $F_t$, such as moving averages, standard deviations, and variability metrics, which can be calculated to enrich the dataset. For instance, a moving average smooths trends over a defined window, while the standard deviation measures variability in the data. These derived features provide complementary information that aligns with the strengths of XGBoost, which excels in handling tabular data and performing feature importance analysis. However, incorporating these features is beyond the scope of this study and is left as a direction for future work.

*4.2 Result Evaluation*

We have run all the three models on the same datasets with the past 7 days historical data as inputs to forecast the current day's temperature. Table 1 compares the performance metrics, RMSE and $R^2$, of three solutions (Transformer, XGBoost, and BTTF) across different training epochs, showing how their predictive performance improves with more training. It also lists the total time costs, covering both training and prediction, for each experiment.

Table 1. Performance Metrics Benchmark Results

|  | **Epochs** | **RMSE** | **R2** | **Time Cost** |
|---|---|---|---|---|
| **Transformer** | 5 | 3.7 | 0.8488 | 10m1s |
| **Transformer** | 100 | 2.5820 | 0.9264 | 3h6m57s |
| **Transformer** | **200** | **2.4635** | **0.9330** | **6h45m59s** |
| **XGBoost (one day)** | 1 | 4.4138 | 0.7886 | 1m42s |
| **XGBoost (time series)** | **1** | **3.9678** | **0.8288** | **1m42s** |
| **BTTF** | 5 | 4.0695 | 0.8192 | 3m21s |
| **BTTF** | 100 | 2.3290 | 0.9407 | 33m21s |
| **BTTF** | **200** | **2.2479** | **0.9448** | **1h3m25s** |

Transformer shows a consistent improvement as the number of training epochs increases. At 5 epochs, the model achieves a moderate RMSE of 3.7 and an R² of 0.8488. Increasing the epochs to 100 yields a marked improvement, with RMSE dropping to 2.5820 and R² rising to 0.9264. At 200 epochs, the trend continues, with RMSE further decreasing to 2.4635 and R² increasing to 0.9330, showing the model's ability to learn and generalize effectively with additional training. This progression highlights the Transformer model's potential for enhanced predictive power as training duration increases.

XGBoost was tested in two configurations: a tabular data setup and a time-series configuration (including 7 days historical data as inputs), both with just 1 epoch. In the tabular setup, it recorded an RMSE of 4.4138 and an R² of 0.7886. The time-series configuration performed better, with a lower RMSE of 3.9678 and an improved R² of 0.8288. Despite the improved results in the time-series setup, XGBoost's performance still lags behind Transformer, which achieve significantly better metrics at comparable or higher training epochs. This underscores XGBoost's limitations with minimal training and highlights the need for more tailored training strategies to enhance its performance.

BTTF shows superior performance and consistently outperforming the Transformer and XGBoost models. At 5 epochs, it matches the Transformer model with an RMSE of 4.0695 and an $R^2$ of 0.8192, showing parity in initial training outcomes. However, as the epochs increase, the BTTF model's advantage becomes evident. At 100 epochs, its RMSE drops to 2.3290, a big improvement from its performance at 5 epochs and a lower value than the Transformer model's RMSE of 2.5820. Its $R^2$ also increases to 0.9407, surpassing the Transformer's 92.64%. At 200 epochs, the BTTF model achieves the best results in the table, with an RMSE of 2.2479 and an $R^2$ of 0.9448. These metrics represent the lowest average error and the highest explained variance, solidifying its status as the best model for predictive accuracy and reliability.

*4.3 Comparative Analysis*

To further understand the reason behind the differences in performance, we did some digging with more insight plotting. Fig. 2, 3, and 4 shows the learning curve, and value distribution charts of Transformer, XGBoost, and the proposed BTTF approach respectively.

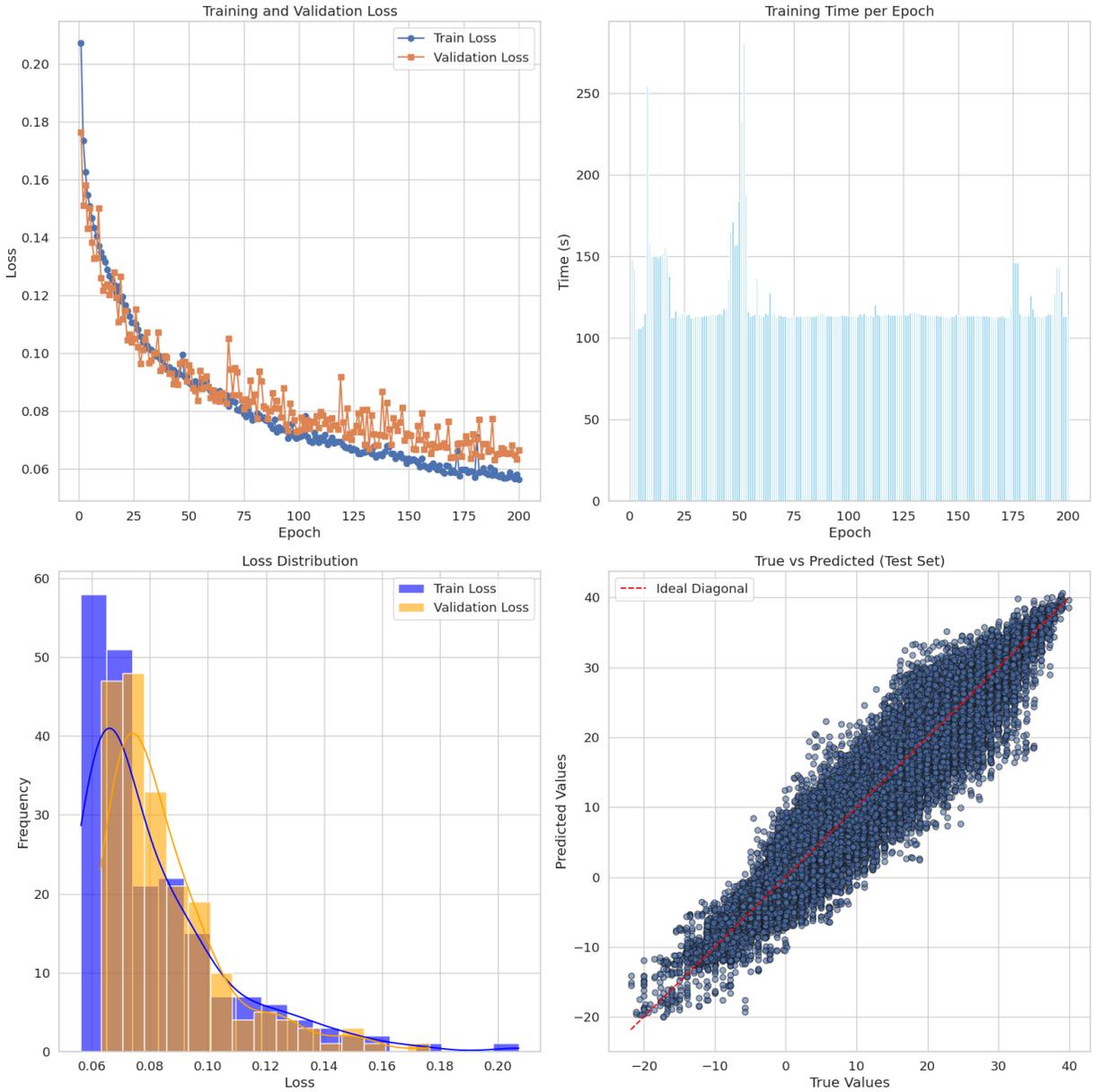

Figure 2. Performance of Transformer Alone (200 epochs)

The top-left plot in Fig. 2 shows the training and validation loss curves over 200 epochs. Both losses exhibit a consistent downward trend, indicating that the model is learning effectively. The gap between the training and validation losses suggests strong generalization. This pattern reflects that the model is not merely memorizing the training data but capturing the underlying relationships within the dataset. The top-right chart shows the training time per epoch, which varies slightly as the training progresses. While earlier epochs demonstrate relatively stable computation times, the later epochs display an increase in duration, particularly around the 20[th] and 50[th] epoch. This could point to adaptive optimization processes or additional computational complexity as the model fine-tunes its parameters near convergence.

The loss distribution, shown in the bottom-left plot, reflects the model's generalization capabilities. The histograms for training and validation losses overlap significantly, with both distributions skewed towards lower loss values. This overlap indicates that the model's performance on the validation data is comparable to its performance on the training data, underscoring its reliability in predicting unseen inputs. The overall distribution suggests that the model effectively minimizes errors across both datasets.

The bottom-right scatter plot compares the predicted values against the true values for the test set. The points cluster tightly along the diagonal line, which represents an ideal predictive scenario where predictions perfectly match the true values. This close alignment indicates that the model has successfully captured the intricate patterns in the data, resulting in minimal residual error. The low spread of points around the diagonal confirms the model's high predictive accuracy and ability to generalize beyond the training and validation datasets.

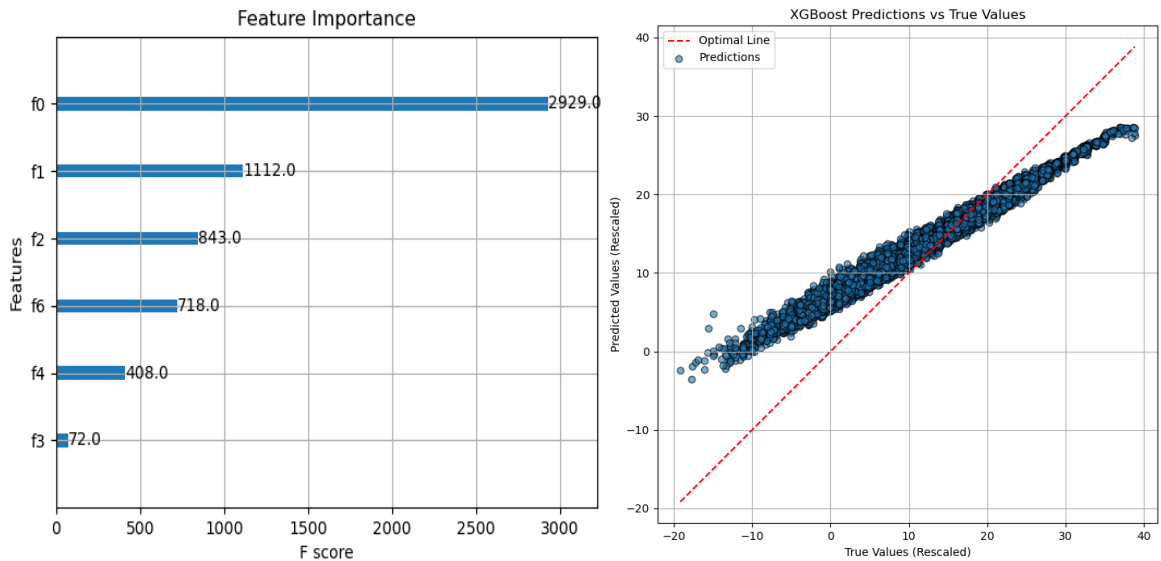

Figure 3. Performance of XGBoost model (f0 - Apparent Temperature, f1 - Humidity, f2 - Wind Speed, f3 - Wind Bearing, f4 - Visibility, f5 - Loud Cover, f6 - Pressure)

Fig. 3 shows the performance of XGBoost, and the feature importance plot (left) highlights that feature "f0" (apparent temperature) significantly contributes to the model's predictions, with an F-score of 2929, followed by "f1" (humidity) and "f2" (wind speed) with scores of 1112.0 and 843.0, respectively. Features like "f6" (pressure), "f3" (wind bearing), and "f4" (visibility) have relatively low importance. The scatter plot (right) illustrates a strong alignment of predicted values with true values along the diagonal optimal line, indicating a high level of accuracy and minimal residual error in the model's predictions. The clustering of points close to the red dashed line shows the model's ability to generalize effectively, capturing the underlying relationships within the data while maintaining robust predictive performance. However, there is a noticeable angle between the distribution and the ideal diagonal line, while the points are tightly aligned rather than scattered. This suggests that XGBoost is better at learning correlations, but it tends to make consistent errors in both the lower and higher value ranges.

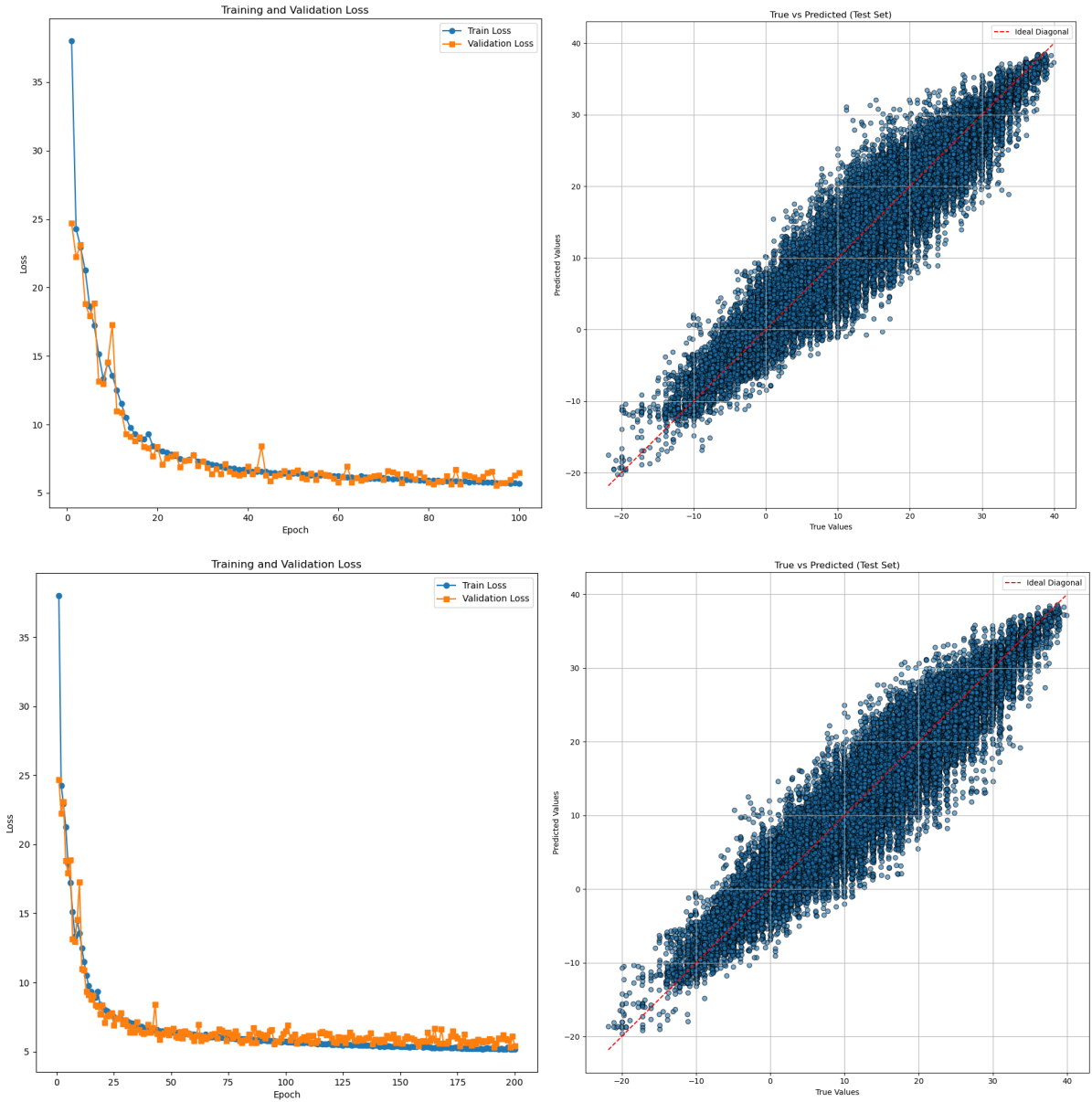

Figure 4. Performance of BTTF framework (100 epochs above; 200 epochs bottom)

Fig. 4 presents the performance of the BTTF approach evaluated at two stages: after 100 epochs and 200 epochs. The first subplot on each row illustrates the training and validation losses across epochs, and the second subplot displays a scatter plot comparing the model's predicted values to the true values on the test set. At epoch 100, the training and validation loss curves demonstrate a rapid decrease during the initial epochs, followed by stabilization around a value of 10. This behavior suggests that the model quickly learns the underlying patterns in the data and maintains good generalization. The close alignment between the training and validation loss curves indicates that the model is not overfitting at this stage. The scatter plot for epoch 100 shows the predicted values closely clustered around the diagonal. The tight distribution of points along this line reflects strong predictive accuracy with minimal residual errors.

At epoch 200, the training and validation loss curves show a similar trend, further stabilizing around the same value of 10. This indicates that additional training does not lead to overfitting but rather contributes to consistent learning. The scatter plot of predicted versus true values at this stage exhibits an even tighter clustering along the diagonal, reflecting a slight improvement in the model's predictive accuracy compared to epoch 100. This improvement suggests that prolonged training has gradually refined the model's ability to capture finer details in the data.

## 5. Discussion

*5.1 Effectiveness of BTTF Framework*

The results demonstrate the advantages of combining the Transformer's forecasting capabilities with XGBoost's adaptive decision-making strengths. The hybrid model shows strong performance across key metrics. The results show it can overcome the limitations of standalone forecasting frameworks, which often struggle either to capture complex, non-linear relationships or to adapt dynamically in real time, as is the case with XGBoost used in isolation. It can continue to improve over more epochs without overfitting easily. The overall learning curves of BTTF on validation set is more stably decreasing comparing to Transformer alone.

Another advantage of the BTTF model is its ability to seamlessly link prediction with present adaptation. In domains such as meteorology, healthcare, or traffic management, it is insufficient to merely predict future conditions. Equally critical is the capacity to make informed decisions in the present based on these forecasts. The hybrid model addresses this need through its feedback loop, which alternates between forecasting future states and adapting present conditions. This ensures that decision-making is continuously informed by robust and dynamic predictions, offering a proactive approach that is particularly advantageous in time-sensitive scenarios where early interventions can mitigate risks and improve outcomes.

XGBoost's feature importance ranking provides a layer of interpretability that is crucial for real-time decision-making applications. By revealing the factors that contribute most significantly to the model's predictions and adaptations, stakeholders are better equipped to prioritize interventions and allocate resources effectively. This transparency not only enhances the usability of the model but also fosters trust in its outputs, which is essential in high-stakes environments such as healthcare and natural disaster response, where decisions have far-reaching consequences.

*5.2 Explanation of the BTTF Performance*

As for the deeper reason driving the better performance of BTTF, it can be attributed to its hybrid architecture, which integrates the strengths of both Transformer-based models and tree-based models like XGBoost. Transformers excel at capturing complex dependencies and patterns in sequential or time-series data through their self-attention mechanisms, enabling the model to learn complicated relationships across long-term data. However, as the model's training increases, it often requires significantly more computational resources, and its learning becomes progressively slower. On the other hand, tree-based models like XGBoost are known for their ability to perform well with tabular data by efficiently handling structured features and capturing non-linear relationships. By combining these two model types, BTTF leverages the powerful pattern recognition capabilities of Transformers while maintaining the efficiency and interpretability of XGBoost. This hybrid approach allows the model to both learn complex sequences and make robust, efficient predictions across different types of data, leading to better generalization, higher predictive accuracy, and more reliable performance, especially when scaling to larger datasets and more epochs.

Meanwhile, BTTF's hybrid design offers an enhanced ability to adapt to varying data complexities. The Transformer component effectively captures intricate temporal or sequential patterns, making it well-suited for data with long-range dependencies, while the XGBoost part can quickly handle tabular features with minimal data preprocessing. By integrating these complementary strengths, BTTF avoids the weaknesses inherent in each model working alone. The Transformer alone can struggle with the need for vast computational resources when trained for long epochs, and XGBoost, although fast and efficient, may not capture long-term dependencies as effectively. The future-visionary aspect of BTTF lies in its ability to continually evolve and refine its predictions, allowing it to dynamically adjust to data trends over time. This fusion of capabilities creates a more versatile, robust model that can handle a wider range of prediction tasks with greater precision, making it better suited for complex, real-world applications.

*5.3 Comparison with Other Forecasting Frameworks*

Another popular framework for forecasting sophisticated situations is the "Chain of Thought" (CoT)[25] framework, which is based on sequential reasoning, where decision-making follows a step-by-step process. Each step in the reasoning chain builds on the previous one, ensuring that logical progression is maintained throughout the task. This approach is particularly useful for complex problems that require detailed, methodical reasoning and where every step needs to be explained in order, such as solving mathematical problems or writing essays. The CoT methodology emphasizes structured, ordered thinking, guiding the system through predefined steps in a coherent manner to reach conclusions.

In contrast, the BTTF methodology is explicitly designed for dynamic environments where conditions change rapidly and immediate action is crucial. Unlike CoT, which focuses on long-term reasoning, BTTF bridges the gap between forecasting (predicting future events) and nowcasting (predicting immediate conditions). This hybrid approach uses forecast outputs not only to gain insights into future trends but also as critical inputs for real-time decisions. By combining these two functions, BTTF enables a system to dynamically adapt to changing conditions, maintaining predictive accuracy while ensuring that immediate actions can be taken in response to the evolving situation. This makes BTTF particularly valuable in areas like meteorology, healthcare, and disaster response, where the need for real-time, adaptive decision-making is paramount.

## 6. Conclusion

This paper presents a BTTF approach demonstrating advancements over one-model-standalone forecasting and adaptation methods, as showcased in the weather forecasting experiment. By combining the strengths of both models, it achieves superior prediction accuracy and enhances real-time decision-making capabilities. The Transformer excels at capturing complex temporal patterns, while XGBoost provides actionable insights for adjustments, making this hybrid framework highly effective for dynamic, real-time applications. The experiments conducted in the weather temperature forecasting use case highlight how this integration results in a more robust forecasting model, better equipped to adapt to changing conditions and provide accurate predictions in environments requiring quick decision-making.

The future work will involve testing additional models to replace the visionary and decision-making components within the BTTF framework, assessing their potential to further improve performance. More efforts will focus on optimizing the hybrid model's operational efficiency, refining the integration of Transformer and XGBoost to streamline processing. Enhancing the framework's ability to make real-time adjustments will help ensure that decisions made today result in better long-term outcomes. These improvements could significantly elevate the overall decision-making capacity of the BTTF framework, particularly in dynamic environments like weather forecasting, where both immediate and future consequences must be carefully considered.